\documentclass[10pt,twocolumn,letterpaper]{article}

\usepackage[pagenumbers]{cvpr} 

\usepackage{graphicx}
\usepackage{amsmath}
\usepackage{amssymb}
\usepackage{booktabs}
\usepackage{color}

\usepackage[accsupp]{axessibility}

\usepackage[capitalize]{cleveref}
\crefname{section}{Sec.}{Secs.}
\Crefname{section}{Section}{Sections}
\Crefname{table}{Table}{Tables}
\crefname{table}{Tab.}{Tabs.}

\definecolor{somegray}{rgb}{0.5, 0.5, 0.5}
\newcommand{\darkgrayed}[1]{\textcolor{somegray}{#1}}
\makeatletter
\newcommand*\titleheader[1]{\gdef\@titleheader{#1}}
\AtBeginDocument{%
  \let\st@red@title\@title
  \def\@title{%
    \vskip-4em
    \bgroup\normalfont\large\centering\@titleheader\par\egroup
    \vskip1.0em\st@red@title}
}
\makeatother

\titleheader{\darkgrayed{This paper has been accepted for publication at the\\
IEEE Conference on Computer Vision and Pattern Recognition Workshops (CVPRW), Vancouver, 2023.
\copyright IEEE}}

\def\cvprPaperID{59} 
\def\confName{CVPR}
\def\confYear{2023}

\title{Neuromorphic Optical Flow and Real-time Implementation with Event Cameras}

\author{Yannick Schnider$^{1,2}$, Stanisław Woźniak$^{1}$, Mathias Gehrig$^{3}$, Jules Lecomte$^{4}$, Axel von Arnim$^{4}$, 
\\Luca Benini$^{2,5}$, Davide Scaramuzza$^{3}$, Angeliki Pantazi$^{1}$
\\[4pt]
$^{1}$IBM Research -- Zurich\ \ 
$^{2}$ETH Zurich\ \ 
$^{3}$University of Zurich\ \ 
$^{4}$fortiss GmbH\ \ 
$^{5}$Università di Bologna
}

\begin{document}

\maketitle


\newcommand{\figOne}[2]{
\begin{figure}[#1]
  \centering
   \includegraphics[width=#2\linewidth]{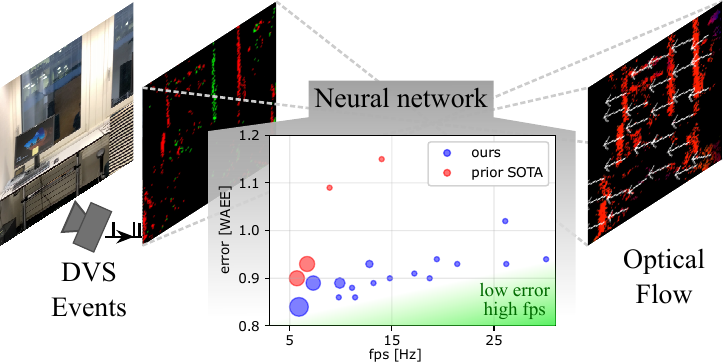}
   \caption{Optical flow estimation from DVS events: We propose a Timelens-based neural network architecture that in comparison with prior art provides lower error and higher real-time framerates.}
   \label{fig:one}
\end{figure}
}

\newcommand{\figArchitecture}[2]{
\begin{figure*}[#1]
  \centering
   \includegraphics[width=#2\linewidth]{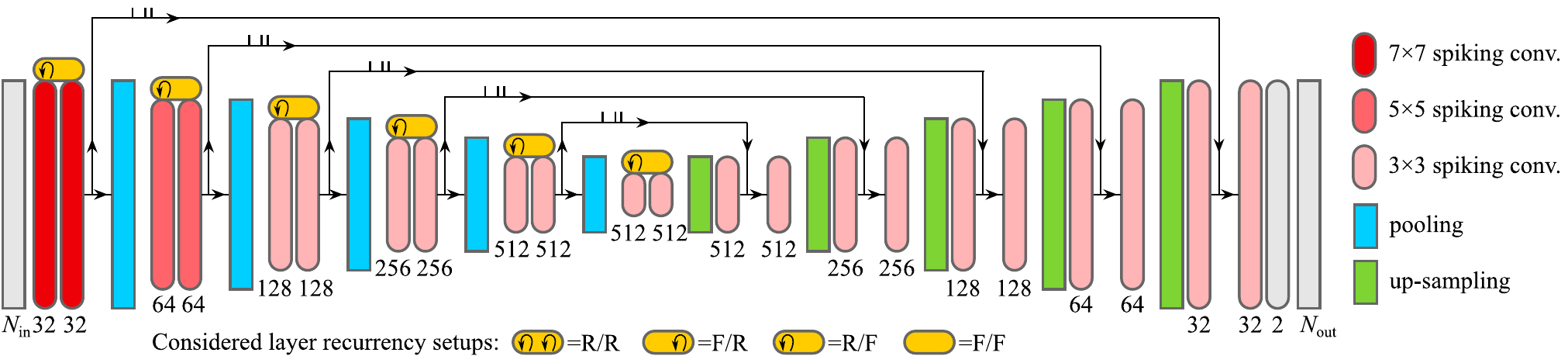}

   \caption{Spiking architecture inspired from Timelens: in the encoding part, we consider different layer-wise reccurency configurations.
%
}
   \label{fig:Architecture}
\end{figure*}
}

\newcommand{\figEncoding}[2]{
\begin{figure}[#1]
  \centering
   \includegraphics[width=#2\linewidth]{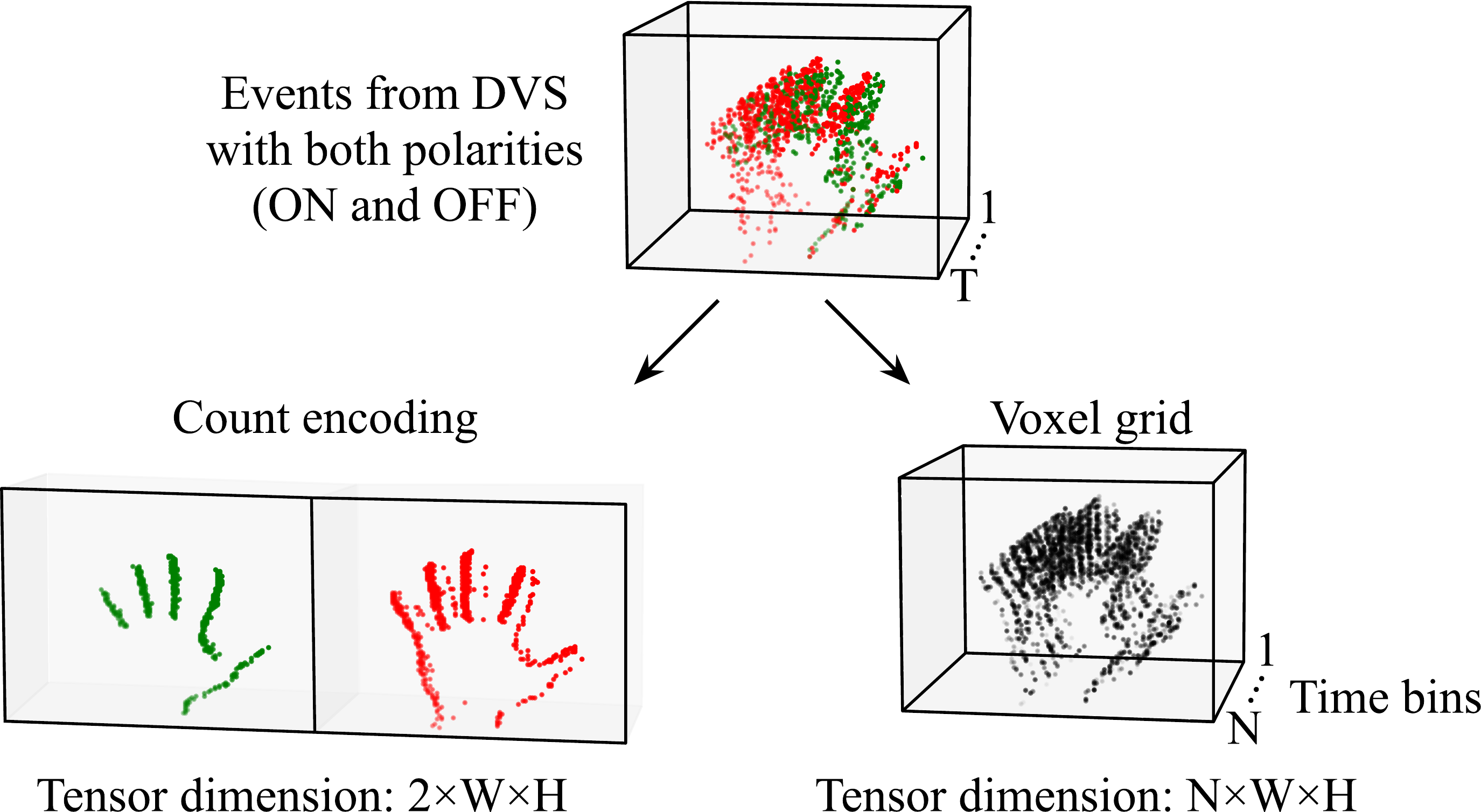}
   \caption{Different input event encodings: Count encoding (per polarity, per pixel) and voxel grid encoding via temporal bi-linear interpolation of combined events into N time bins.}
   \label{fig:encoding}
\end{figure}
}

\newcommand{\figChannelReduction}[2]{
\begin{figure}[#1]
  \centering
   \includegraphics[width=#2\linewidth]{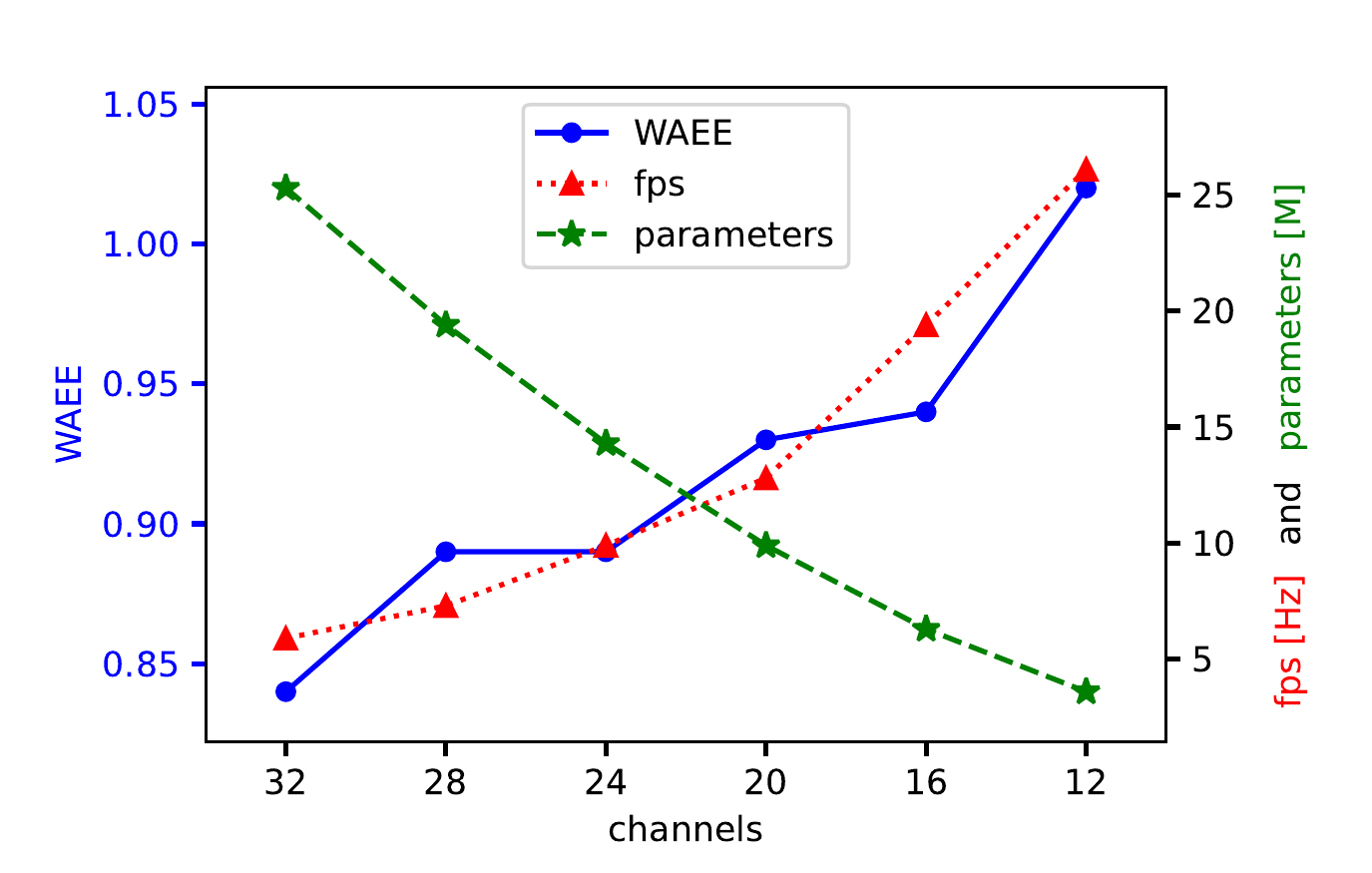}
   \caption{Impact of convolutional channels count: WAEE, network parameters (in millions [M]), and inference frequency (in frames per second [fps]) for SNN-Timelens with 5 stages.}
   \label{fig:channel_reduction}
\end{figure}
}

\newcommand{\figReduction}[2]{
\begin{figure}[#1]
  \centering
   \includegraphics[width=#2\linewidth]{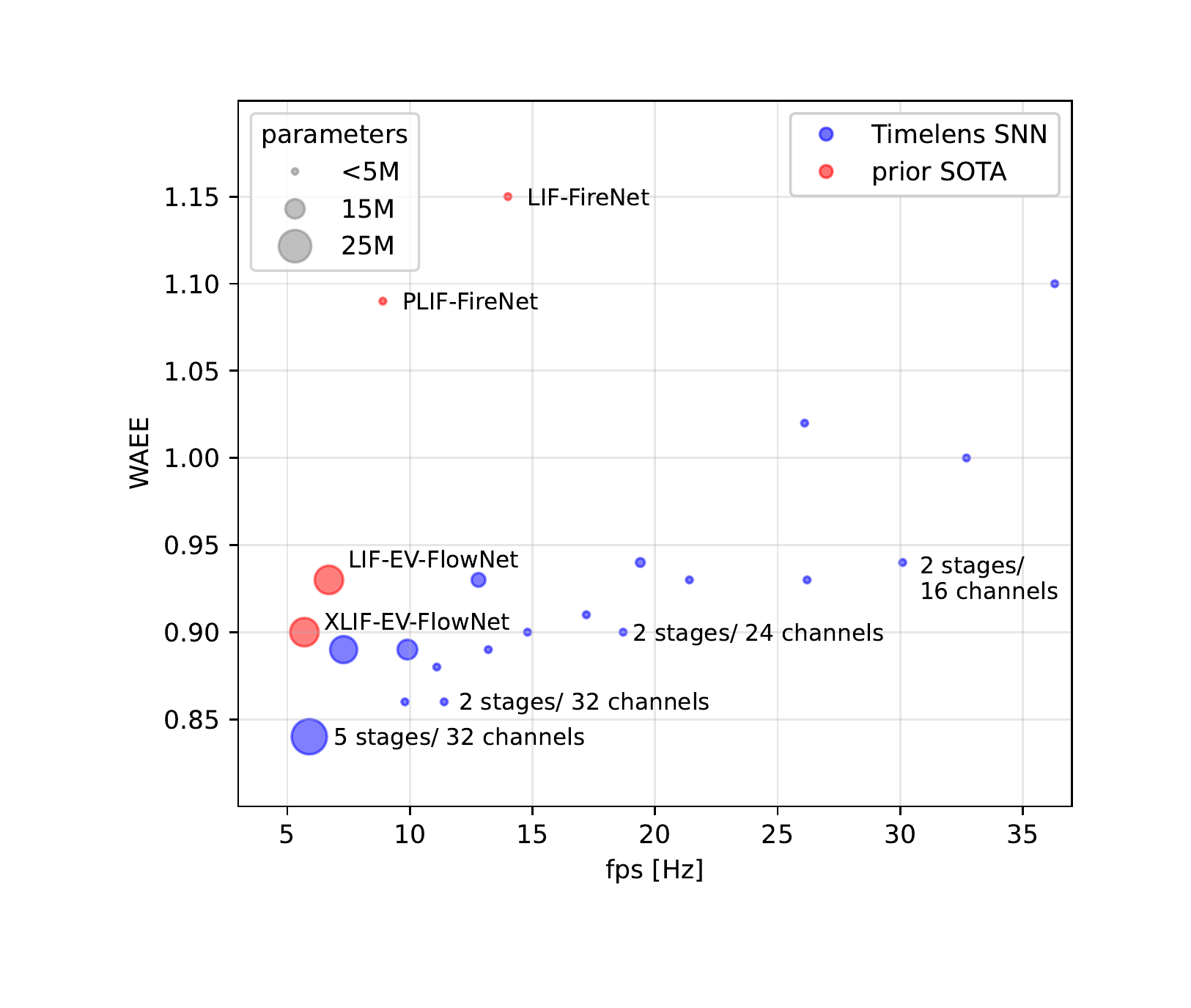}
   \caption{Model reduction results: SNN-Timelens  compared with state-of-the-art (SOTA) in our CPU setup. WAEE plotted versus frames per second (fps); circle size indicates model size. For readability, only selected SNN-Timelens from Table \ref{tab:channels} are labeled.}
   \label{fig:reduction}
\end{figure}
}
\newcommand{\figSpikingAnalysis}[2]{
\begin{figure*}[#1]
  \centering
   \includegraphics[width=#2\linewidth]{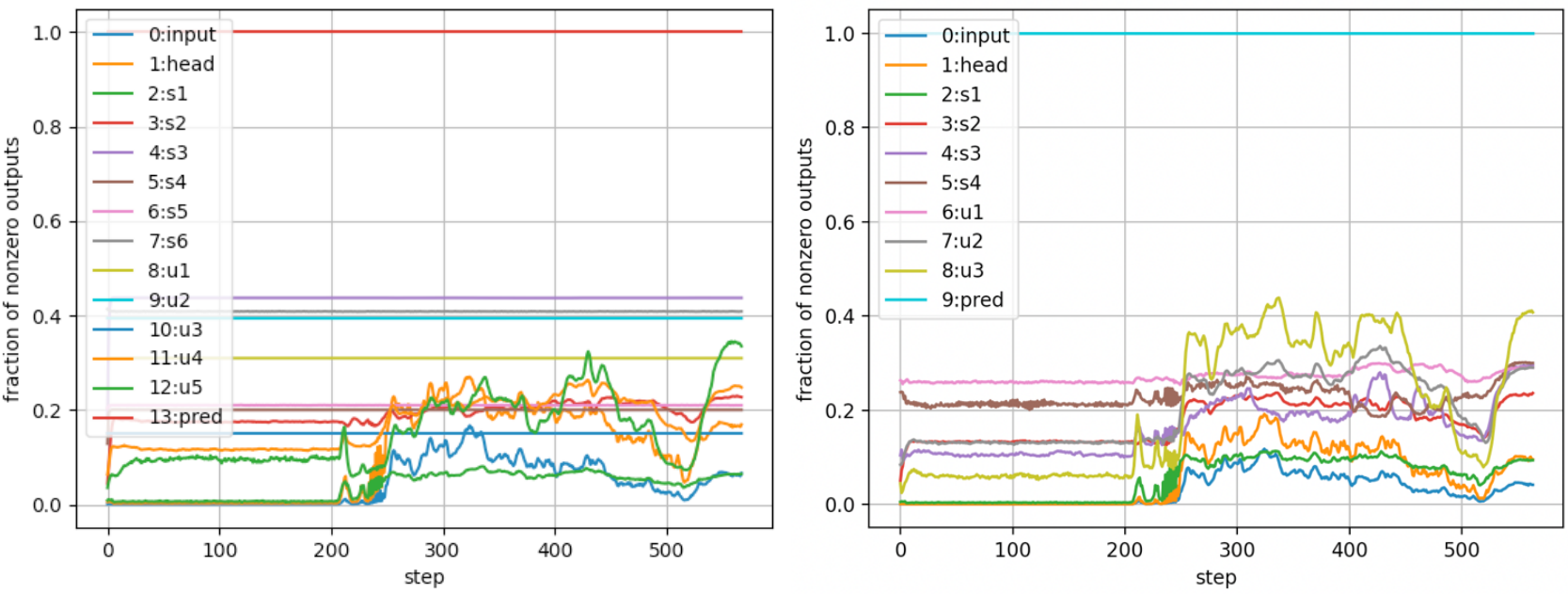}

   \caption{Spiking activity for an MVSEC test sequence: The fraction of spiking neurons is registered for all layers from input to prediction layer. Left: An architecture with 5 encoding and decoding blocks, Right: a reduced architecture with 3 encoding and decoding blocks.}
   \label{fig:SpikingAnalysis}
\end{figure*}
}

\newcommand{\figDemoHand}[2]{
\begin{figure*}[#1]
  \centering
   \includegraphics[width=#2\linewidth]{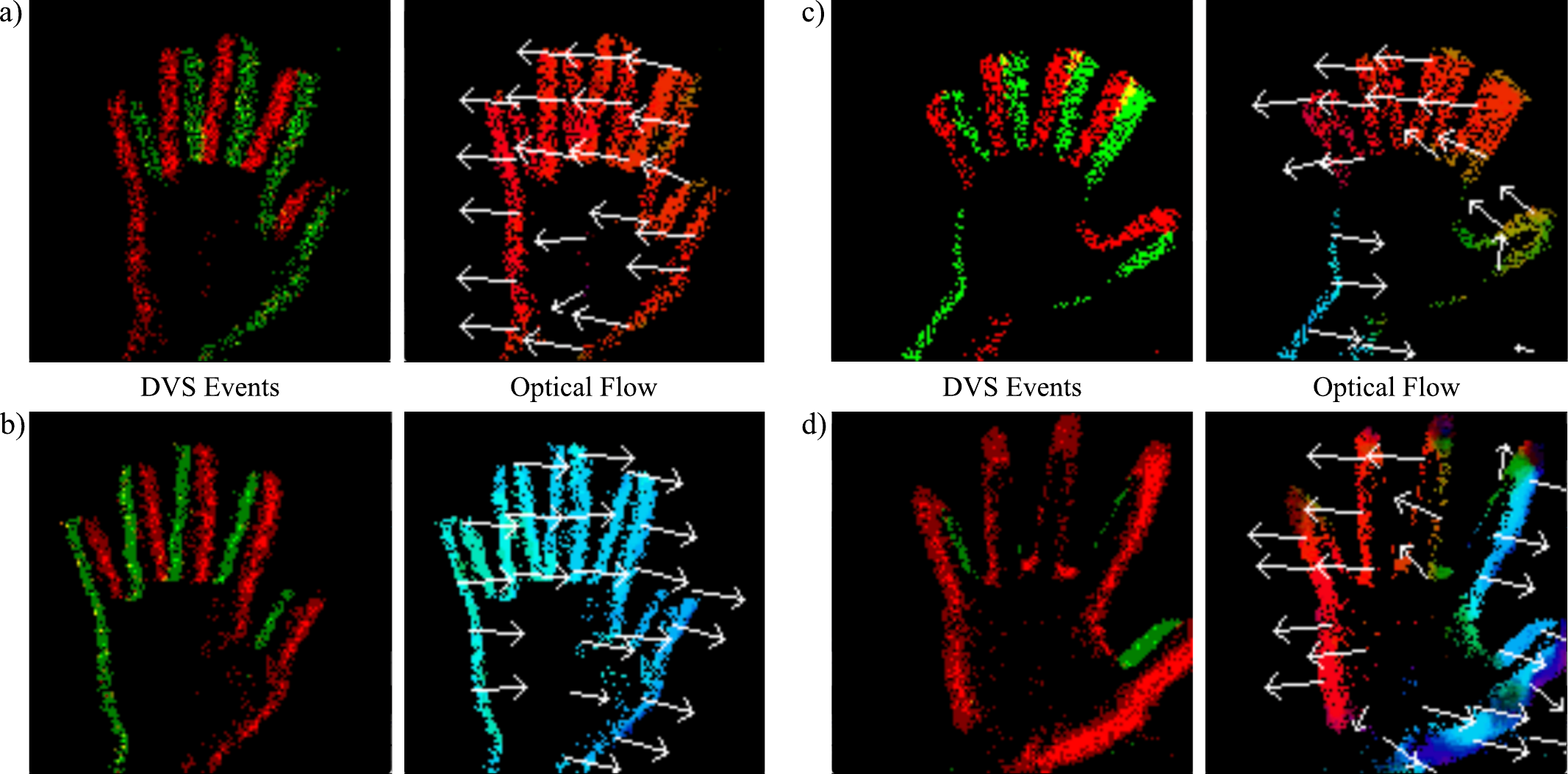}

   \caption{Real-time predictions: DVS events aggregated over one aggregation window and the corresponding optical flows from reduced SNN-Timelens (0.32M) applied for different movements of a hand: (a) to the right, (b) to the left, (c) rotation, (d) approaching the camera.}
   \label{fig:DemoHand}
\end{figure*}
}

\newcommand{\figColorWheel}[2]{
\begin{figure}[#1]
  \centering
   \includegraphics[width=#2\linewidth]{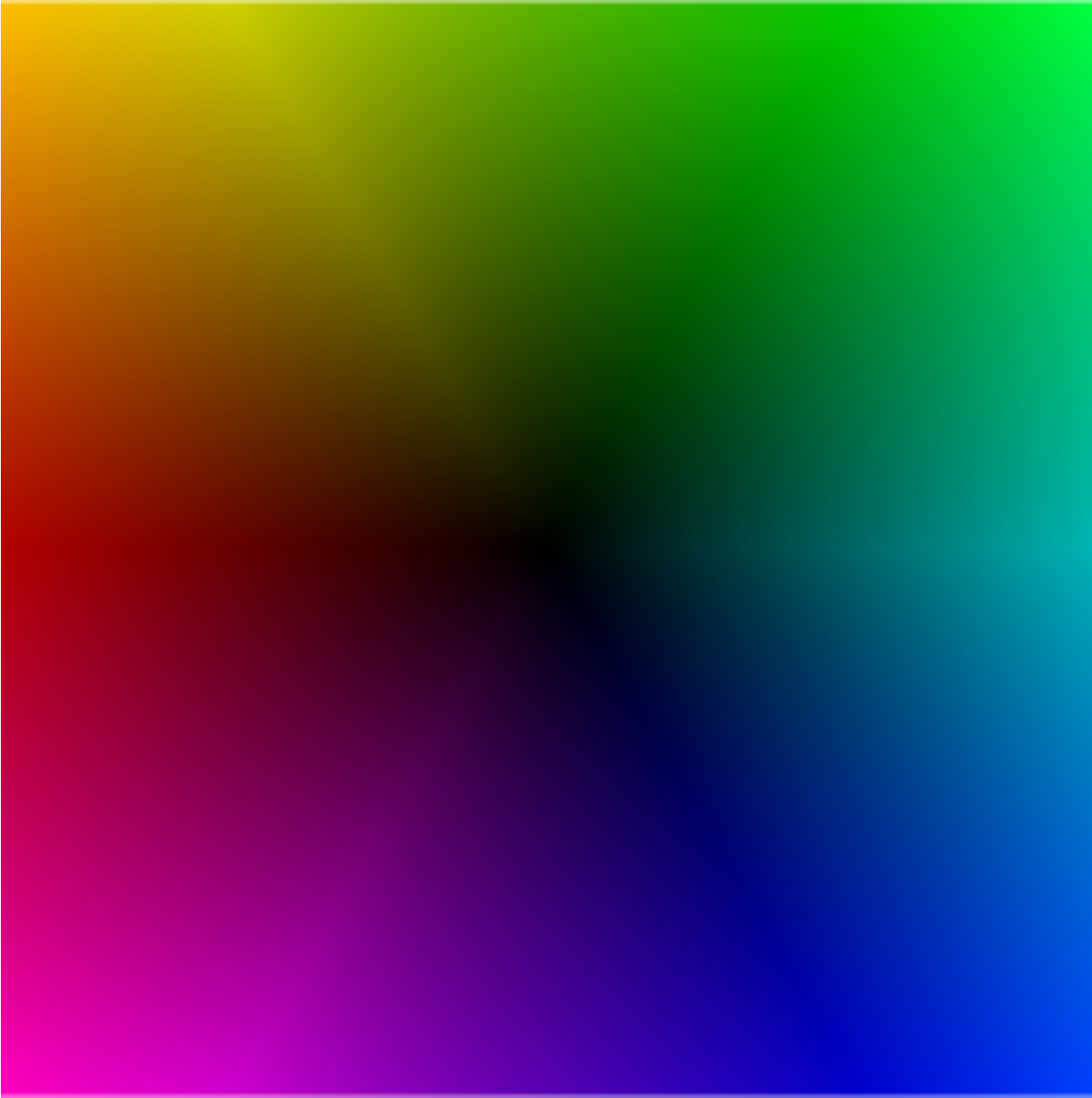}
   \caption{Colour wheel used for angular encoding of the optical flow.}
   \label{fig:colorwheel}
\end{figure}
}

\newcommand{\figComplex}[2]{
\begin{figure*}[#1]
  \centering
  \begin{subfigure}{0.68\linewidth}
    \fbox{\rule{0pt}{2in} \rule{.9\linewidth}{0pt}}
    \caption{An example of a subfigure.}
    \label{fig:short-a}
  \end{subfigure}
  \hfill
  \begin{subfigure}{0.28\linewidth}
    \fbox{\rule{0pt}{2in} \rule{.9\linewidth}{0pt}}
    \caption{Another example of a subfigure.}
    \label{fig:short-b}
  \end{subfigure}
  \caption{Example of a short caption, which should be centered.}
  \label{fig:short}
\end{figure*}
}

\newcommand{\tabRecurrency}[1]{
\begin{table}[#1]
  \centering
  \begin{tabular}{@{}lccccc@{}}
    \toprule[1.5pt]
     &  \multicolumn{2}{c}{SNN-Timelens} & \phantom{}& \multicolumn{2}{c}{sSNU-Timelens}\\  
    \cmidrule{2-3} \cmidrule{5-6} 
    $\text{dt} = 1$ & WAEE & $\overline{\text{\%}}\textsubscript{\text{Outlier}}$ && WAEE & $\overline{\text{\%}}\textsubscript{\text{Outlier}}$ \\
    \midrule
    R/F & \textbf{0.84} & \textbf{4.10} &&   \underline{0.77}& \underline{4.23}\\
    F/R & \underline{0.85} & 4.36 && 1.11 & 8.44\\
    R/R & 0.89 & \underline{4.26} &&   \textbf{0.73}& \textbf{3.89}\\
    F/F & 1.12& 5.89 &&   1.18& 9.26\\
    \midrule
    \\ 
    $\text{dt} = 4$ \\
    \midrule
    R/F &  \textbf{0.84} & \textbf{32.88} &&  \underline{0.74}& \underline{27.20}\\
    F/R & \underline{0.86} & \underline{34.23} && 1.13& 44.92\\
    R/R & 0.90& 35.81 &&  \textbf{0.71}& \textbf{25.66}\\
    F/F &  1.15 & 52.36 &&   1.19& 48.34\\
    \bottomrule[1.5pt]
  \end{tabular}
  \caption{Effects of layer recurrency placement on WAEE (the lower, the better $\downarrow$) and~$\overline{\text{\%}}\textsubscript{\text{Outlier}}$($\downarrow$) in the encoding blocks. Best scores are in bold, while runner-ups are underlined.
  }
  \label{tab:recurrency}
\end{table}
}
\newcommand{\tabMultiLayerLoss}[1]{
\begin{table}[#1]
  \centering
  \begin{tabular}{@{}lccccc@{}}
    \toprule[1.5pt]
    Recurrency &  \multicolumn{2}{c}{$\text{dt} = 1$ } & \phantom{}& \multicolumn{2}{c}{$\text{dt} = 4$ }\\  
    \cmidrule{2-3} \cmidrule{5-6} 
    & WAEE & increase && WAEE & increase \\
    \midrule
    R/F multi & 0.92 & 9.52 \% && 0.92 & 9.52\% \\
    F/R multi & 0.92 & 8.24 \% &&  0.92 & 6.98\% \\
    R/R multi & 0.94 & 5.62 \% &&  0.95& 5.56\% \\
    F/F multi & 1.33& 18.75 \% && 1.39 & 16.0\% \\
    \bottomrule[1.5pt]
  \end{tabular}
  \caption{Effects of multi-layer loss function on intermediate up-sampled flow predictions for different layer recurrency placement in the encoder. WAEE($\downarrow$) and its relative increases with regard to the last layer loss in Table \ref{tab:recurrency} for SNN-Timelens.}
  \label{tab:multi_layer_loss}
\end{table}
}
\newcommand{\tabChannels}[1]{
\begin{table}[#1]
  \centering
  \begin{tabular}{@{}lcccccc@{}}
    \toprule[1.5pt]
    \# channels &32 & 28 &24 &20 &16 &12\\
    \midrule
    5 stages\\ \midrule
    WAEE & 0.84 & 0.89 &0.89 & 0.93 & 0.94 & 1.02\\
    parameters & 25.3 & 19.4 & 14.3 & 9.9 & 6.3 &3.6 \\
    frequency & 5.9 & 7.3  & 9.9& 12.8 &  19.4& 26.1 \\
    \midrule
    3 stages\\ \midrule
    WAEE & 0.86 & 0.88 &0.90 & 0.91 & 0.93 & 1.00\\
    parameters & 1.75 & 1.34 & 0.98 & 0.68 & 0.44 &0.25 \\
    frequency & 9.8 & 11.1  & 14.8& 17.2 &  26.2& 32.7 \\
     \midrule
    2 stages\\ \midrule
    WAEE & 0.86 & 0.89 &0.90 & 0.93 & 0.94 & 1.10\\
    parameters & 0.57 & 0.44 & 0.32 & 0.23 & 0.15 &0.08 \\
    frequency & 11.4 & 13.2 & 18.7& 21.4 &  30.1& 36.3 \\
    \bottomrule[1.5pt]
  \end{tabular}
  \caption{Impact of convolutional channels count: WAEE, number of network parameters (in millions [M]), and inference frequency (in frames per second [fps]) of our Timelens-based SNNs for 5, 3 and 2 stages (encoding/decoding blocks).}
  \label{tab:channels}
\end{table}
}

\newcommand{\tabEncoders}[1]{
\begin{table}[#1]
  \centering
  \begin{tabular}{@{}lcccc@{}}
    \toprule[1.5pt]
    \# stages &5 & 4 &3 &2 \\
    \midrule
    WAEE & 0.84 & 0.87 &0.86 &0.86\\
    parameters & 25.3 & 6.5 & 1.75 &0.57 \\
    frequency &5.9 & 8.5  & 9.8 & 12.4 \\
    \bottomrule[1.5pt]
  \end{tabular}
  \caption{Impact of stages count (encoders/decoders): WAEE, number of network parameters (in millions [M]), and inference speed (in frames per second [fps]) of our Timelens-based SNNs.}
  \label{tab:encoders}
\end{table}
}

\newcommand{\tabReduction}[1]{
\begin{table}[#1]
  \centering
  \begin{tabular}{@{}lcccc@{}}
    \toprule[1.5pt]
    \# model &A & B &C &D \\
    \midrule
    WAEE & 0.84 & 0.87 &0.86 &0.86\\
    parameters & 25.3 & 6.5 & 1.75 &0.57 \\
    frequency &5.9 & 8.5  & 9.8 & 12.4 \\
    \bottomrule[1.5pt]
  \end{tabular}
  \caption{TODO: CURRENTY DUMMY DATA. Model reduction results for our Timelens-based SNNs. (Preferrably a plot rather than a table).}
  \label{tab:reduction}
\end{table}
}

\newcommand{\tabBig}[1]{
\newcommand{\pout}[0]{\%\textsubscript{Out.}} 
\newcommand{\ra}[1]
{\renewcommand{\arraystretch}{#1}}
\begin{table*}\centering
\ra{1.3}
\begin{tabular}{@{}lrrrcrrrcrrrcrrr@{}}\toprule[1.5pt]
& \multicolumn{2}{c}{outdoor\textunderscore day1} & \phantom{}& \multicolumn{2}{c}{indoor\textunderscore flying1} &
\phantom{} & \multicolumn{2}{c}{indoor\textunderscore flying2} &
\phantom{} & \multicolumn{2}{c}{indoor\textunderscore flying3} &\phantom{} & \multicolumn{2}{c}{overall}\\ \cmidrule{2-3} \cmidrule{5-6} \cmidrule{8-9} \cmidrule{11-12} \cmidrule{14-15}

$\text{dt}=1$& AEE & \pout && AEE & \pout  && AEE & \pout && AEE & \pout & \multicolumn{2}{r}{WAEE} &$\overline{\text{\%}}\textsubscript{\text{Out.}}$\\
\midrule
LIF-EV-FlowNet\cite{TU_Delft} & 0.53 & 0.33  && 0.71 & 1.41 && 1.44 & 12.75 && 1.16 & 9.11 && 0.93 & 5.90\\
XLIF-EV-FlowNet\cite{TU_Delft} & 0.45 & \textbf{0.16}  && 0.73 & \underline{0.92} && 1.45 & 12.18 && 1.17 & 8.35 && 0.90 & 5.40\\ 
LIF-FireNet\cite{TU_Delft} & 0.57 & 0.40  && 0.98 & 2.48 && 1.77 & 16.40 && 1.50 & 12.81 && 1.15 & 8.02\\
PLIF-FireNet\cite{TU_Delft} & 0.56 & 0.38  && 0.90 & 1.93 && 1.67 & 14.47 && 1.41 & 11.17 && 1.10 & 7.00\\
our SNN-Timelens & \underline{0.44} & 0.18 && \underline{0.70} & \textbf{0.79} && \underline{1.30} & \underline{9.41} && \underline{1.05} & \underline{6.00} && \underline{0.84} & \underline{4.10}\\ 
our SNUo-Timelens & \textbf{0.39} & \underline{0.17} && \textbf{0.64} & 0.96 && \textbf{1.17} & \textbf{7.71} && \textbf{0.96} & \textbf{4.92} && \textbf{0.76} & \textbf{3.44} \\\midrule 
EV-FlowNet\cite{TU_Delft} & \underline{0.47} & \underline{0.25} && \underline{0.60} & \textbf{0.51} && \textbf{1.17} & \textbf{8.06} && \textbf{0.93} & \underline{5.64} && \underline{0.78} & \textbf{3.61} \\
RNN-EV-FlowNet\cite{TU_Delft} & 0.56 & 1.09 && 0.62 & 0.97 && 1.20 & 8.82 && \textbf{0.93} & \textbf{5.51} && 0.83 &4.10 \\ 
our sSNU-Timelens & \textbf{0.36} & \textbf{0.10} && \textbf{0.58} & \underline{0.56} && \underline{1.19} & \underline{8.78} && \underline{0.96} & 6.11 && \textbf{0.73} & \underline{3.89}\\\midrule   
\\ 
$\text{dt}=4$\\ \midrule
LIF-EV-FlowNet\cite{TU_Delft} & 2.02 & 18.91  && 2.63 & 29.55 && 4.93 & 51.10 && 3.88 & 41.49 && 0.92 & 35.26\\
XLIF-EV-FlowNet\cite{TU_Delft} & 1.67 & 12.69  && 2.72 & 31.69 && 4.93 & 51.36 && 3.91 & 42.52 && 0.89 & 34.57\\ 
LIF-FireNet\cite{TU_Delft} & 2.12 & 21.00  && 3.72 & 48.27 && 6.27 & 64.16 && 5.23 & 58.43 && 1.17 & 47.97\\
PLIF-FireNet\cite{TU_Delft} & 2.11 & 20.64  && 3.44 & 44.02 && 5.94 & 64.02 && 4.98 & 57.53 && 1.11 & 46.55\\
our SNN-Timelens & \underline{1.65} & \underline{11.03} && \underline{2.61} & \underline{29.40} && \underline{4.50} & \underline{50.87} && \underline{3.58} & \underline{40.22} && \underline{0.84} & \underline{32.88}\\  

our SNUo-Timelens & \textbf{1.44} & \textbf{8.98} && \textbf{2.36} & \textbf{24.18} && \textbf{3.98} & \textbf{44.71} && \textbf{3.25} & \textbf{36.01} && \textbf{0.75} & \textbf{28.47} \\
\midrule
EV-FlowNet\cite{TU_Delft} & \underline{1.69} & \underline{12.50} && \underline{2.16} & \underline{21.51} && \textbf{3.90} & \textbf{40.72} && \textbf{3.00} & \textbf{29.60} && \underline{0.74} & \underline{26.08}\\
RNN-EV-FlowNet\cite{TU_Delft} & 1.91 &16.39 && 2.23 &22.10 && 4.01 & 41.74 &&\underline{3.07} & \underline{30.87} && 0.78 & 27.78 \\ 
our sSNU-Timelens  & \textbf{1.34} & \textbf{7.99} && \textbf{2.15} & \textbf{20.92} && \underline{3.97} & \underline{41.31} && 3.17 & 32.44 && \textbf{0.71} & \textbf{25.67}\\ \bottomrule[1.5pt]   
\end{tabular}
\caption{Evaluation on the MVSEC dataset for comparable models trained on UZH-FPV Drone Racing Dataset: AEE (the lower, the better $\downarrow$), the percentage of outliers $\text{\%}\textsubscript{\text{Out.}}$($\downarrow$) per sequence, and the overall WAEE($\downarrow$) as defined in Eq.~\ref{eq:WAEE} as well as the average percentage of outliers $\overline{\text{\%}}\textsubscript{\text{Out.}}$($\downarrow$). Best scores are in bold, while runner-ups are underlined. Horizontal lines delimit the spiking and the non-spiking models.}
\label{tab:results}
\end{table*}}


\begin{abstract}   
Optical flow provides information on relative motion that is an important component in many computer vision pipelines. Neural networks provide high accuracy optical flow, yet their complexity is often prohibitive for application at the edge or in robots, where efficiency and latency play crucial role. To address this challenge, we build on the latest developments in event-based vision and spiking neural networks. We propose a new network architecture, inspired by Timelens, that improves the state-of-the-art self-supervised optical flow accuracy when operated both in spiking and non-spiking mode. To implement a real-time pipeline with a physical event camera, we propose a methodology for principled model simplification based on activity and latency analysis. We demonstrate high speed optical flow prediction with almost two orders of magnitude reduced complexity while maintaining the accuracy, opening the path for real-time deployments.
\end{abstract}
\\[-0.3cm]
\noindent
\textbf{Video}: \url{https://youtu.be/jDGDxKabj0o}

\section{Introduction}
\label{sec:intro}

Optical flow is defined as an apparent motion of objects, edges and surfaces in a visual scene registered by the observer. It is caused by the relative motion between the observer and the scene and does not distinguish between actual motion in the visual scene and change in the observer's pose. 
The applications of optical flow in the field of computer science include motion estimation and video compression  \cite{Motion_Brox2011, Agustsson_2020_CVPR}. In machine perception as well as in robotics, optical flow is used for both object detection and tracking \cite{detection_robotics, objectDetectionAslani2013, trackingSatelliteDu}, robot navigation and even for control of micro air vehicles \cite{SLAM_Zhang, optical_flow_control_UAV, motionFlow_Scaramuzza}. 

\figOne{t}{1.0}
\vspace{-5pt}

Deployment of optical flow in real-time robotic scenarios requires low-latency processing and energy efficiency. Existing algorithms usually calculate optical flow at discrete rates based on frames obtained from conventional cameras \cite{Dosovitskiy_FlowNet2015}. Neuromorphic dynamic vision sensors (DVS) operate similarly to the eye's retina by providing a continuous stream of events representing brightness changes rather than absolute measurements at fixed time intervals \cite{gallego_event-based_2022}. Since optical flow computation relies on regions and time instants where brightness changes, DVS represents a viable alternative for fast optical flow prediction, demonstrated in recent works \cite{brebion_2021,Shiba_2022}.
Moreover, the sparsity of the events can be exploited by spiking neural networks (SNN) as opposed to artificial neural networks (ANNs). The advantage of SNNs deployed on neuromorphic hardware is low latency and energy efficiency coming from sparse computations \cite{SNN_Energy_neuro}.   

Recently, researchers have presented an approach to produce sparse optical flow based on event data with SNNs \cite{TU_Delft}. However, there is a disconnect between large-scale architecture modelling and real-time deployments in efficient hardware. Here, we present a novel approach of a Timelens\cite{time_lens}-like network for sparse optical flow predictions. Apart from surpassing the optical flow baseline in terms of the average endpoint error (AEE) \cite{TU_Delft}, we also address the deployment aspect through systematic model reduction and demonstrate real-time operation with a physical DVS camera, as schematically illustrated in Fig.~\ref{fig:one}.

\vspace{10pt}
\textbf{This paper makes the following contributions:}

\begin{enumerate}
    \item We design an optical flow architecture inspired by the Timelens architecture, enriched with spiking neurons operating with DVS-based inputs.
    \item We surpass the state-of-the-art self-supervised optical flow performance for SNNs and ANNs on the MVSEC dataset \cite{Zhu_2018_mvsec} for event-based vision, reducing the prediction error by 6.1\% in spiking, 15.6\% in analog-valued spiking, and 5.5\% in non-spiking mode.
    \item We propose a principled methodology, involving activity and latency analysis, for reducing the network size to fit into realistic real-time hardware constraints.
    \item We demonstrate model reduction from 20.4M to 0.32M parameters with 0\% penalty in error with regard to the prior art \cite{TU_Delft}, enabling real-time operation with DVS inputs.
\end{enumerate}

\figArchitecture{}{1.0}

\section{Related Work}

\subsection{Deep learning of SNNs}
\label{sec:SNU}
In recent years, SNN popularity in machine learning has been increasing owing to research advancements that enabled easy modelling and training in deep learning frameworks \cite{stan_nature, neftci_surrogate_2019}.
Beyond the standard Leaky Integrate-and-Fire (LIF) model, an even wider variety of neuro-inspired spiking models has been explored. In particular, a framework around so-called Spiking Neural Unit (SNU) includes the plain SNU with LIF dynamics and typical axo-dendritic synapses, as well as its variants that model further biological aspects, such as axo-axonic and axo-somatic synapses in SNUo and SNUa, respectively. These variants demonstrated improvements in large-scale speech recognition models~\cite{bohnstingl_speech_2022}.
In the context of optical flow, modifications of a LIF implementation were also proposed and called ALIF, XLIF, and PLIF \cite{TU_Delft}. 


\subsection{Architectures for optical flow}
Successful training of neural networks relies on a proper loss definition, where historically supervised losses were used \cite{teed2020raft, Dosovitskiy_FlowNet2015, gehrig2021raft}. Due to challenges with obtaining a large number of high-quality labels, it is beneficial to reformulate the training in terms of a self-supervised loss \cite{Zhu_EVFlownet, Liu_2019_SelFlow}. In \cite{TU_Delft}, the optical flow prediction task was posed as a self-supervised contrast maximization problem.
This training approach can be applied to popular network architectures for optical flow prediction that include EV-FlowNet \cite{Zhu_EVFlownet} and FireNet \cite{Scheerlinck_2020_FireNet}. 
State-of-the-art SNN implementations are based on their adaptation to inputs from event-based cameras and the operation with spiking neurons \cite{TU_Delft}.


\subsection{Timelens}
The Timelens architecture \cite{time_lens} was proposed in the context of event-based video frame interpolation. The design itself has been inspired from the hourglass network with skip connections for frame-based video interpolation -- a problem posed initially in \cite{super_slomo}. A peculiarity of the network architecture is the reduction of the spatial dimensionality in the encoding part using a pooling operator rather than exploiting strided convolutions. Another feature is the bigger kernel sizes for the initial two convolutions compared with the rest of the encoding/decoding blocks.

\section{Network model}
We propose an architecture for prediction of optical flow based on SNNs receiving an event stream from DVS.
Design choices, such as spatial down- and up-sampling, channel dimensions, kernel sizes and skip connections, are inspired by the Timelens network \cite{time_lens}. Our network is reformulated as an SNN by incorporating spiking spatial convolutions featuring stateful neural cells and layer recurrency.
An overall architectural diagram is presented in Fig.~\ref{fig:Architecture} and the details are described in the following subsections.

\subsection{Neuron models}

First, we implement an SNN using state equations that describe the common neuroscientific LIF model in a form trainable within the realm of deep learning \cite{neftci_surrogate_2019, stan_nature, TU_Delft}:
\begin{align}
\label{eq:SNU}
s_t &= (1-d)(W x_t + H y_{t-1}) + d s_{t-1} (1-y_{t-1}) \\
\label{eq:SNUh}
y_t &= h(s_t - v_{\text{th}}),
\end{align}
where $s_t$ is the state -- membrane potential voltage of the neuron, $W$ and $H$ are the input and optional recurrent weights, respectively, $d$ is membrane potential decay factor, $y_t$ is the output, $v_{\text{th}}$ is a firing voltage threshold, and $h$ is the step activation function. 
The model is trainable with backpropagation-through-time assuming a smooth derivative of $\operatorname{arctanspike}(x, a) = 1/(1 + a\cdot x^2)$ for $h$, with $a=10$. Trainable parameters include $W$, $H$, $d$ and $v_{\text{th}}$.
This neuronal model is our main focus and we will quantify architectures using it with the SNN prefix.

Secondly, we consider a more advanced biologically-inspired extension of the basic LIF -- the so-called SNUo unit, which models the concept of axo-axonic synapses that enrich neuronal connectivity by modulating the neuronal outputs \cite{bohnstingl_speech_2022}. From implementation perspective, this leads to emission of sparse analog-valued spikes, or graded spikes in the nomenclature of Intel's Loihi 2 implementation \cite{loihi2_2021}.
The equations of SNUo are \cite{bohnstingl_speech_2022}:
\begin{align}
s_t &= g(W x_t + H y_{t-1} + d s_{t-1} (1-\Tilde{y}_{t-1})) \\
\Tilde{y}_t &= h(s_t - v_{\text{th}}) \\
y_t &= \Tilde{y}_t \phantom{i} o(W_o x_t + H_o y_{t-1} + b_o),
\label{eq:SNUO}
\end{align}
where $\Tilde{y}_t$ is the unmodulated neuron output used for resetting the membrane potential, $y_t$ is the modulated output propagating to downstream units, and $g$ is an additional activation function that we set to leaky $\operatorname{ReLU}$ with a leak of 0.1. $W_o$, $H_o$ and $b_o$ are additional trainable parameters. We use the $sigmoid$ function as activation $o$ for output modulation in Eq.~\ref{eq:SNUO} to mimic the inhibitory character of the axo-axonic synapses as suggested in \cite{bohnstingl_speech_2022}. We will quantify networks using this approach with the SNUo prefix.

Lastly, the benefits of neuromorphic internal dynamics were demonstrated also in the non-spiking mode: 
by operating with real-values in the so-called soft SNU (sSNU) approach \cite{stan_nature}.
The idea is to replace the step activation function $h$ with $sigmoid$ function in Eq.~\ref{eq:SNUh}.
As sSNU operates by continuously outputting real values, we will benchmark it against non-spiking baselines. We will quantify networks using this unit with the sSNU prefix.

\subsection{Network structure}
\label{sec:Architecture}
Spiking convolutions work similarly to conventional convolutions found in ANNs except for the neural dynamics applied to their outputs. The computed per pixel, per channel outputs of the 2D convolutions serve as input currents ($Wx_t$ in Eq.~\ref{eq:SNU}) for the spiking neural units. 
Simultaneously, layer-wise recurrency ($Hy_{t-1}$ in Eq.~\ref{eq:SNU}) is an additional feature to capture temporal dependencies that is not always considered in SNN modelling. We explicitly mention whenever we do include this term.

The network structure is illustrated in Fig.~\ref{fig:Architecture}. Its first stage comprises two spiking 2D convolutions expanding the $N_{\text{in}}$ input channels to 32 output channels featuring $7 \times 7$ kernels. While the spatial dimension is retained for the spiking convolution by using stride 1 and appropriate zero padding, spatial down-sampling is performed afterwards using 2D average pooling with kernel size $2\times 2$. The remaining encoding parts of the network are five similar encoding blocks consisting of two spiking convolutions followed by pooling operators. The kernel sizes are $3 \times 3$, except for the first encoding block ($5 \times 5$). For each encoding block the number of output channels is doubled while the spatial resolution is halved. For the two spiking convolutions in each encoding block, we consider all combinations of layer-wise recurrency, as marked in Fig.~\ref{fig:Architecture}.

For decoding, five identical decoding blocks are used. Each consists of 2D bilinear up-sampling by a factor of 2, followed by two spiking convolutions. The number of output channels gets halved with each decoding block and the convolutional kernel sizes are $3 \times 3$. Skip connections between each encoder/decoder pair of the same resolution provide values which are concatenated channel-wise before the second spiking convolution in each decoding block. 

To obtain continuous optical flow values, the final layer is a $1 \times 1$ convolution with $tanh$ activation. This layer reduces 32 base channels to $N_{\text{out}}=2$ channels representing the optical flow components $u$ and $v$ that correspond to horizontal and vertical optical flow magnitudes, respectively.

\figEncoding{b}{1.0}

\subsection{Input coding}
\label{sec:input_encoding}
The DVS event stream contains events of the form:
\begin{equation}
e_i = (x_i, y_i, t_i, p_i)    
\end{equation}
where $x_i$ and $y_i$ represent the pixel coordinates, $t_i$ the timestamp and $p_i$ the ON/OFF polarity of the event. 
Different encoding strategies have been proposed to process the raw event stream data prior to feeding it into a neural network. Commonly used input coding techniques are the count encoding \cite{count_encoding} and the voxel grid encoding \cite{voxel_encoding_IWE}, depicted in Fig.~\ref{fig:encoding}. The count encoding loses the temporal information of single events within the aggregation window. Events get accumulated per pixel and per polarity for the entire window width. On the other hand, a voxel-based representation discretizes the time span of the aggregation window and uses temporal bi-linear interpolation to populate the bins with events. Polarity is not treated as a separate channel, but negative OFF events (-1) and positive ON events (+1) are summed in a single channel.

For our spiking architectures, we opted for the voxel grid input coding. The number of discrete time bins is an additional hyperparameter. Choosing the number of bins too high yields overly sparse inputs while for a low number of bins the encoding collapses to a count representation with a single channel. In the latter case, positive and negative events can annihilate each other leading to information loss. 
For our spiking network, performance peaked at six time bins ($N_{\text{in}}=6$).
However, when operating in the non-spiking mode of sSNU, the count encoding with separate ON/OFF channels ($N_{\text{in}}=2$) performed better, so we use it for sSNU-based networks.
To ensure a fair comparison, the aggregation window width is fixed and the set of encoded events is therefore the same for both encoding approaches. 

\subsection{Training setup}
All models are trained in a self-supervised fashion on the UZH-FPV Drone Racing Dataset \cite{drone_racing_data}, using the approach and configurations from \cite{TU_Delft}. Specifically, contrast maximization loss is applied there to compensate the motion and predict optical flow from the input events. The loss is:
\begin{equation}
    \mathcal{L} = \mathcal{L}_{\text{contrast}}(t_{\text{ref}}^{\text{fw}}) +  \mathcal{L}_{\text{contrast}}(t_{\text{ref}}^{\text{bw}}) +  \lambda \mathcal{L}_{\text{smoothing}}
\end{equation}
where contrast maximization is performed in a forward ($t_{\text{ref}}^{\text{fw}}$) as well as a backward ($t_{\text{ref}}^{\text{bw}}$) fashion w.r.t. the current reference time instance $t_{\text{ref}}$.
$\mathcal{L}_{\text{smoothing}}$ is a Charbonnier smoothness prior \cite{Charbonnier} proposed in \cite{Zhu_EVFlownet, voxel_encoding_IWE} and $\lambda=0.001$ is a balancing constant. Truncated back-propagation through time (TBPTT) is performed after every 10 forward passes. 

In the original approach \cite{TU_Delft}, the loss included different spatial resolutions of the optical flow maps. 
We analogously extended our architecture with 2D convolutions with $tanh$ activation to produce optical flow predictions of different resolutions at each decoding block. These intermediate optical flow maps are then up-sampled to the initial spatial dimension using nearest neighbour interpolation for the loss computation. Simultaneously, they are concatenated to the input channels of the subsequent decoding blocks.

However, in contrast to the prior work, 
we also considered an architecture with the loss applied only to the last output layer's prediction. This approach is simpler, faster and  turned out to be beneficial for our  architecture. 

\tabRecurrency{b}

\tabMultiLayerLoss{b}

\tabBig{}

\section{Simulation results}
The quantitative performance and generalization abilities of the trained models (self-supervised on the UZH-FPV Drone Racing Dataset) are evaluated on the MVSEC dataset \cite{Zhu_2018_mvsec} following the comparison approach from \cite{TU_Delft}. The predicted sparse optical flow is compared against the ground truth optical flow provided by \cite{Zhu_EVFlownet}. The ground truth labels are available at timestamps corresponding with conventional camera's frames and quantify the optical flow over one ($\text{dt} = 1$) or four  ($\text{dt} = 4$) frames. 

The well-established average end point error (AEE) in pixels is used to evaluate the four sequences of the dataset: outdoor\_day1 ($\text{od1}$), indoor\_flying1 ($\text{if1}$), indoor\_flying2 ($\text{if2}$), indoor\_flying3 ($\text{if3}$). For easier comparability, we introduce a weighted average endpoint error (WAEE) to combine the four metrics into a single scalar value: 
\begin{equation}
\begin{aligned}
  \text{WAEE} =& (\frac{\text{AEE}_{\text{od1}}}{w_{\text{od1}}} + \frac{\text{AEE}_{\text{if1}}}{w_{\text{if1}}} +\frac{\text{AEE}_{\text{if2}}}{w_{\text{if2}}}  +\frac{\text{AEE}_{\text{if3}}}{w_{\text{if3}}} )/4,
  \label{eq:WAEE}
\end{aligned}
\end{equation}
where the four weights 
are based on the average AEE of the best-performing spiking architectures
of the prior art \cite{TU_Delft} -- 
see Supplementary Note 1 for the values for each dt setting.

Using the WAEE metric, we explored different configurations of the layer recurrency in the convolutional blocks, visualized in Fig.~\ref{fig:Architecture}.
As each block comprises two spiking convolutions, there are four different combinations of recurrent (R) and feed-forward (F) convolutions: R/F, F/R, R/R and F/F. Table \ref{tab:recurrency} reports the results in terms of WAEE and the average percentage of outliers $\overline{\text{\%}}\textsubscript{\text{Outlier}}$. When operating in the spiking mode, having one convolution with layer recurrency per block is favourable. In particular, best performance is achieved with recurrent layers in the first convolution (R/F). On the contrary, in the context of non-spiking mode of sSNU, double layer recurrency (R/R) is beneficial. We use these best configurations for the final models.

We also evaluated an implementation of multi-resolution loss, described in the training section.
For both settings of $\text{dt} = 1$ and $\text{dt} = 4$,
the reported WAEE values in Table \ref{tab:multi_layer_loss} demonstrate that using the simpler setup of the loss applied only at the last layer is preferred for our architecture. 
A possible interpretation of the observed deterioration is that the multi-layer loss function trains the deeper decoders to encode down-sampled optical flow rather than to develop higher-level features. 
Furthermore, such a formulation is inconsistent with the ultimate task of the network, which is predicting high-resolution optical flow at the last layer rather than outputting the flow predictions  at multiple intermediate stages. Imposing a loss only on the last layer, omits this restriction.
We use this approach for all our models.

\figSpikingAnalysis{h}{0.91}

The resulting AEEs, WAEEs and outlier percentages ($\text{AEE}>3$ pixels) for our Timelens-based architecture with spiking (SNN), analog-valued spiking (SNUo) and non-spiking (sSNU) units are reported in Table \ref{tab:results}. 
Our model is compared with the  state-of-the-art spiking and non-spiking architectures trained in the identical self-supervised setting \cite{TU_Delft}.
For an extended comparison with EV-FlowNet \cite{voxel_encoding_IWE,Zhu_EVFlownet} and Hybrid-EV-FlowNet \cite{spike_flownet} 
that use different training datasets and setups, see Supplementary Note 2.

For spiking neural networks, our SNN-Timelens surpasses the performance of the LIF- and XLIF-EV-FlowNet by 9.7\%, 6.7\% with regard to WAEE and lowers the percentage of outliers $\overline{\text{\%}}\textsubscript{\text{Outlier}}$ by 30.5\%, 24.1\% for $\text{dt}=1$, respectively. As the improvement over the XLIF-EV-FlowNet is 5.6\% for $\text{dt}=4$, the average prediction error is reduced by 6.1\%. 
Table \ref{tab:results} shows that our SNNs are not only better on average, but outperform the comparable state-of-the-art on each MVSEC sequence for $\text{dt} = 1$ and $\text{dt} = 4$. 

Operating with analog-valued spikes, SNUo-Timelens achieves a further substantial reduction in WAEE: 18.3\% and 15.6\% vs. LIF- and XLIF-EV-FlowNet for $\text{dt}=1$, respectively. Since the improvement over XLIF-EV-FlowNet for $\text{dt}=4$ is the same, an average decrease is 15.6\%. It is also remarkable that SNUo-Timelens demonstrates on-par performance with the best non-spiking self-supervised prior-art for comparable training configurations.

Lastly, the best performing model is the sSNU-Timelens incorporating neuromorphic dynamics into the non-spiking mode of operation. Despite featuring 8.5\% less parameters than the best baseline EV-FlowNet (32.9M), our sSNU-TimeLens (30.1M) outperforms it with regard to WAEE for both $\text{dt} = 1$, $\text{dt} = 4$ by 6.4\%, 4.1\%, respectively. The average improvement over the state-of-the-art for comparable non-spiking models therefore equals 5.5\%. 


\section{Model reduction for real-time operation}
The state-of-the-art models listed in Table~\ref{tab:results} involve tens of millions of parameters and are executed on high-end GPUs.
To close the gap between large-scale architecture modelling and real-time deployment, model reduction is required. 
We propose a principled approach for model reduction that includes analysis of the network activity and of the relationships between the number of parameters and inference speed at different stages of the architecture.

We focus our exploration on the SNN-Timelens that could benefit the most from efficient implementation on SNN chips, such as TrueNorth \cite{TrueNorth}, Loihi \cite{davies_loihi:_2018} or Kraken's SNE \cite{di2022kraken}, that support the LIF equations used in the SNU. If support for analog-valued spikes increases, as in Loihi 2, the SNUo-Timelens architecture could become appealing.

\subsection{Spiking activity analysis}
\label{sec:spiking_analysis}
A spiking activity analysis has been conducted to obtain potential information about the importance of different network building blocks. For a test sequence of the MVSEC dataset, the fraction of neurons that produced spikes was registered for each network layer: input layer, initial convolutional layers, encoding layers \texttt{s[i]}, decoding layers \texttt{u[i]} and the final prediction layer. Fig.~\ref{fig:SpikingAnalysis} shows the spiking activity in SNN-Timelens architecture with 5 encoding/decoding blocks (left) compared with a network reduced to 3 encoding/decoding blocks (right). In the following we will refer to the number of encoding/decoding blocks as the number of stages of the Timelens model. 

In general, the fraction of non-zero outputs, which corresponds to the fraction of neurons that spike, is almost constant until time step 210. At this time step, the drone in the DVS recording lifts off and the incoming events actually come from movement rather than static noise. The spiking activity for all layers fluctuates between 0 and 0.5 when optical flow is predicted due to the actual movement. It has to be noted that the activity of the last layer is 1.0 for all times since the final prediction layer does not feature a $step$ but rather a continuous $tanh$ activation function. 

For the bigger model comprising 5 stages the fraction of non-zero outputs does not vary at all for the deep encoding \texttt{s3} - \texttt{s5} and decoding \texttt{u1} - \texttt{u3} layers. However, evaluation of the gradients indicates that the weights get updated during training. The question therefore arises whether these deep layers are crucial for the overall model performance. Reducing the number of stages from 5 to 3 shows indeed almost on par performance, only 2.6 \% WAEE drop on MVSEC, while the spiking activity varies for all layers. The smaller model features only 1.75M parameters, which is 14.5 times less than the initial SNN architecture with 25.35M. The constant spiking activity for the layers can be interpreted as a quasi-identity mapping between early encoding and late decoding layers. Thus, dropping these layers tends to have a minor effect on the network capabilities.


\figChannelReduction{b}{1.0}

\tabChannels{b}

\figReduction{b}{1.0}

\figDemoHand{h}{0.92}

\subsection{Network profiling}

In deep CNNs there is no simple linear relationship between the number of parameters and the inference latency. Therefore, we profiled the contributions of the components of the model to assess how the number of stages (encoding/decoding blocks) and the size of convolution impacts the inference frequency in frames per second (fps).
Model performance is monitored throughout the process to find a balance between speed and quality of the predicted optical flow. The fps values are calculated from timings of 100 forward passes on 128$\times$128 DVS inputs using Pytorch code executed on a single core of Intel Core i7 2.6GHz CPU.

\textbf{Reducing channels.~}Network profiling has revealed that the first convolution and the first encoding block are particularly costly in terms of computations. On one hand this is due to large spatial input dimension, on the other hand it is influenced by the big convolutional kernels ($7 \times 7$ and $5\times 5$). Nevertheless, decreasing the number of output channels effectively reduces the computational costs. 
Fig. \ref{fig:channel_reduction} illustrates a trade-off between the number of channels and performance in terms of WAEE and fps for the SNN-Timelens model with 5 stages. Note the non-linear relationship between convolutional channels and network parameters.

\textbf{Reducing stages.~}The spiking analysis showed that less than 5 stages, e.g. 3 stages, are sufficient to obtain reasonable optical flow predictions. Table \ref{tab:channels} extends the analysis, reporting the WAEE ($\text{dt} = 1$), the number of network parameters and model inference frequency for different number of channels and stages. Comparing the WAEE between 5 and 2 stages, we observe minor performance degradation: 0.84 versus 0.86. 
The 2-stage model comes with 44.4 times less parameters and increases the evaluation frequency by 93.2\%. For further speedup, the number of channels of the 2-stage SNN-Timelens model can be decreased at the cost of degrading performance in terms of WAEE.


\section{Model reduction results}


The comparison of our architecture before and after reduction is presented for a set of selected configurations in Fig.~\ref{fig:reduction}.
While our initial SNN-Timelens featured 5 stages with 32 channels and used 25.35M parameters, our model after reduction features only 2 stages with 32 channels and 0.57M parameters, thus reducing the number of trainable parameters by a factor of 44.4. It involves a trade-off in terms of WAEE performance that degrades by just 2.6\% (0.84 versus 0.86). Remarkably, it still remains better than the prior state-of-the-art large models of LIF- and XLIF-EV-FlowNet (20.4M) with WAEE 0.90 and 0.93, respectively. Note that to match prior art performance (WAEE 0.90), our SNN-Timelens needs only 2 stages with 24 channels (0.32M), featuring 63.75 times less parameters.


\subsection{Qualitative results}

For qualitative performance assessment and validation of the generalization ability of the last proposed network with 2 stages and 24 channels, a complete real-time pipeline was implemented to process the event stream of a DVS128 camera from iniVation AG. 
Fig.~\ref{fig:DemoHand} shows the optical flow predictions of different hand movements in front of the DVS. While color-coding is used to encode the optical flow, additionally a sparse arrow grid is superimposed to the optical flow for instant intuitive validation of the predictions. Arrow angles and magnitudes represent direction and magnitude of the biggest optical flow within a local $10 \times 10$ neighborhood of pixels, respectively.

The predicted flow in Fig.~\ref{fig:DemoHand} looks  reasonable and coincides with the expected dislocations caused by the moving hand. Linear motion is correctly captured (left plots) and the model generalizes well to more challenging scenarios such as rotating or approaching hand (right plots).

\section{Conclusion}

In this work we proposed a neuromorphic solution for optical flow estimation comprising an event camera combined with a Timelens-inspired architecture.
We demonstrated SNN, SNUo and sSNU versions of our architecture, operating with different biologically inspired neuron models.
By tuning the architectural design, the event encoding, the placement of recurrent connections, and the loss function formulation, we improved the performance in comparison with prior art models on the MVSEC dataset. 
Our architecture surpassed both SNN and ANN baselines when operating in spiking and real-valued modes, respectively. Remarkably, when operating with analog-valued spikes, it demonstrated performance comparable to the ANN baseline. 
Furthermore, a principled model reduction approach was proposed to meet realistic real-time hardware constraints. 
Our SNN-Timelens model reduced to 0.32M parameters achieves WAEE on-par with the state-of-the-art while decreasing the number of parameters by almost two orders of magnitude. Finally, a real-time pipeline was demonstrated with a physical DVS camera. 
Future work includes deployment of the proposed architecture on a neuromorphic SNN chip to further decrease the latency and increase energy efficiency.




\section{Acknowledgements}
\label{sec:ack}
The research was carried out in collaboration with the IBM and fortiss Center for AI (C4AI). We thank its team for discussions and assistance.
The research at fortiss was supported by the HBP Neurorobotics Platform funded through the European Union’s Horizon 2020 Framework Program for Research and Innovation under the Specific Grant Agreements No.\,945539 (Human Brain Project SGA3).


{\small
\bibliographystyle{ieee_fullname}
\bibliography{egbib}

\begin{thebibliography}{10}\itemsep=-1pt

\bibitem{Agustsson_2020_CVPR}
Eirikur Agustsson, David Minnen, Nick Johnston, Johannes Balle, Sung~Jin Hwang,
  and George Toderici.
\newblock Scale-space flow for end-to-end optimized video compression.
\newblock In {\em Proceedings of the IEEE/CVF Conference on Computer Vision and
  Pattern Recognition (CVPR)}, Jun 2020.

\bibitem{TrueNorth}
Filipp Akopyan, Jun Sawada, Andrew Cassidy, Rodrigo Alvarez-Icaza, John Arthur,
  Paul Merolla, Nabil Imam, Yutaka Nakamura, Pallab Datta, Gi-Joon Nam, Brian
  Taba, Michael Beakes, Bernard Brezzo, Jente~B. Kuang, Rajit Manohar,
  William~P. Risk, Bryan Jackson, and Dharmendra~S. Modha.
\newblock {TrueNorth}: Design and tool flow of a 65 {mW} 1 million neuron
  programmable neurosynaptic chip.
\newblock {\em IEEE Transactions on Computer-Aided Design of Integrated
  Circuits and Systems}, 34(10):1537--1557, 2015.

\bibitem{objectDetectionAslani2013}
Sepehr Aslani and Homayoun Mahdavi-Nasab.
\newblock Optical flow based moving object detection and tracking for traffic
  surveillance.
\newblock {\em International Journal of Electrical, Computer, Energetic,
  Electronic and Communication Engineering}, 7(9):1252--1256, 2013.

\bibitem{SNN_Energy_neuro}
Adarsha Balaji, Anup Das, Yuefeng Wu, Khanh Huynh, Francesco~G. Dell’Anna,
  Giacomo Indiveri, Jeffrey~L. Krichmar, Nikil~D. Dutt, Siebren Schaafsma, and
  Francky Catthoor.
\newblock Mapping spiking neural networks to neuromorphic hardware.
\newblock {\em IEEE Transactions on Very Large Scale Integration (VLSI)
  Systems}, 28(1):76--86, 2020.

\bibitem{bohnstingl_speech_2022}
Thomas Bohnstingl, Ayush Garg, Stanislaw Wozniak, George Saon, Evangelos
  Eleftheriou, and Angeliki Pantazi.
\newblock Speech {Recognition} {Using} {Biologically}-{Inspired} {Neural}
  {Networks}.
\newblock In {\em {ICASSP} 2022 - 2022 {IEEE} {International} {Conference} on
  {Acoustics}, {Speech} and {Signal} {Processing} ({ICASSP})}, pages
  6992--6996, Singapore, Singapore, May 2022. IEEE.

\bibitem{brebion_2021}
Vincent Brebion, Julien Moreau, and Franck Davoine.
\newblock Real-time optical flow for vehicular perception with low- and
  high-resolution event cameras.
\newblock {\em IEEE Transactions on Intelligent Transportation Systems},
  PP:1--13, 12 2021.

\bibitem{Motion_Brox2011}
Thomas Brox and Jitendra Malik.
\newblock Large displacement optical flow: Descriptor matching in variational
  motion estimation.
\newblock {\em IEEE Transactions on Pattern Analysis and Machine Intelligence},
  33(3):500--513, 2011.

\bibitem{Charbonnier}
P. Charbonnier, L. Blanc-Feraud, G. Aubert, and M. Barlaud.
\newblock Two deterministic half-quadratic regularization algorithms for
  computed imaging.
\newblock In {\em Proceedings of 1st International Conference on Image
  Processing}, volume~2, pages 168,169,170,171,172, Los Alamitos, CA, USA, Nov
  1994. IEEE Computer Society.

\bibitem{optical_flow_control_UAV}
Guido Croon, Christophe De~Wagter, and Tobias Seidl.
\newblock Enhancing optical-flow-based control by learning visual appearance
  cues for flying robots.
\newblock {\em Nature Machine Intelligence}, 3:33--41, Jan 2021.

\bibitem{davies_loihi:_2018}
Mike Davies, Narayan Srinivasa, Tsung-Han Lin, Gautham Chinya, Yongqiang Cao,
  Sri~Harsha Choday, Georgios Dimou, Prasad Joshi, Nabil Imam, Shweta Jain,
  Yuyun Liao, Chit-Kwan Lin, Andrew Lines, Ruokun Liu, Deepak Mathaikutty,
  Steven McCoy, Arnab Paul, Jonathan Tse, Guruguhanathan Venkataramanan,
  Yi-Hsin Weng, Andreas Wild, Yoonseok Yang, and Hong Wang.
\newblock Loihi: {A} {Neuromorphic} {Manycore} {Processor} with {On}-{Chip}
  {Learning}.
\newblock {\em IEEE Micro}, 38(1):82--99, Jan 2018.

\bibitem{drone_racing_data}
Jeffrey Delmerico, Titus Cieslewski, Henri Rebecq, Matthias Faessler, and
  Davide Scaramuzza.
\newblock Are we ready for autonomous drone racing? the {UZH-FPV} {Drone}
  {Racing} {Dataset}.
\newblock In {\em 2019 International Conference on Robotics and Automation
  (ICRA)}, pages 6713--6719, 2019.

\bibitem{di2022kraken}
Alfio Di~Mauro, Moritz Scherer, Davide Rossi, and Luca Benini.
\newblock Kraken: {A} {Direct} {Event}/{Frame}-{Based} {Multi}-sensor {Fusion}
  {SoC} for {Ultra}-{Efficient} {Visual} {Processing} in {Nano}-{UAVs}.
\newblock In {\em 2022 {IEEE} {Hot} {Chips} 34 {Symposium} ({HCS})}, pages
  1--19, Cupertino, CA, USA, Aug 2022. IEEE.

\bibitem{Dosovitskiy_FlowNet2015}
Alexey Dosovitskiy, Philipp Fischer, Eddy Ilg, Philip Hausser, Caner Hazirbas,
  Vladimir Golkov, Patrick van~der Smagt, Daniel Cremers, and Thomas Brox.
\newblock Flownet: Learning optical flow with convolutional networks.
\newblock In {\em Proceedings of the IEEE International Conference on Computer
  Vision (ICCV)}, Dec 2015.

\bibitem{trackingSatelliteDu}
Bo Du, Shihan Cai, and Chen Wu.
\newblock Object tracking in satellite videos based on a multiframe optical
  flow tracker.
\newblock {\em IEEE Journal of Selected Topics in Applied Earth Observations
  and Remote Sensing}, 12(8):3043--3055, 2019.

\bibitem{gallego_event-based_2022}
Guillermo Gallego, Tobi Delbruck, Garrick Orchard, Chiara Bartolozzi, Brian
  Taba, Andrea Censi, Stefan Leutenegger, Andrew~J. Davison, Jorg Conradt,
  Kostas Daniilidis, and Davide Scaramuzza.
\newblock Event-{Based} {Vision}: {A} {Survey}.
\newblock {\em IEEE Transactions on Pattern Analysis and Machine Intelligence},
  44(1):154--180, Jan 2022.

\bibitem{gehrig2021raft}
Mathias Gehrig, Mario Millh{\"a}usler, Daniel Gehrig, and Davide Scaramuzza.
\newblock E-{RAFT}: Dense optical flow from event cameras.
\newblock In {\em 2021 International Conference on 3D Vision (3DV)}, pages
  197--206. IEEE, 2021.

\bibitem{motionFlow_Scaramuzza}
Volker Grabe, Heinrich~H B{\"u}lthoff, Davide Scaramuzza, and Paolo~Robuffo
  Giordano.
\newblock Nonlinear ego-motion estimation from optical flow for online control
  of a quadrotor {UAV}.
\newblock {\em The International Journal of Robotics Research},
  34(8):1114--1135, 2015.

\bibitem{TU_Delft}
Jesse Hagenaars, Federico Paredes-Vall\'es, and Guido de Croon.
\newblock Self-supervised learning of event-based optical flow with spiking
  neural networks.
\newblock {\em Advances in Neural Information Processing Systems}, 34, 2021.

\bibitem{super_slomo}
Huaizu Jiang, Deqing Sun, Varun Jampani, Ming-Hsuan Yang, Erik Learned-Miller,
  and Jan Kautz.
\newblock Super {SloMo}: High quality estimation of multiple intermediate
  frames for video interpolation.
\newblock In {\em Proceedings of the IEEE Conference on Computer Vision and
  Pattern Recognition (CVPR)}, June 2018.

\bibitem{spike_flownet}
Chankyu Lee, Adarsh Kosta, Alex Zhu, Kenneth Chaney, Kostas Daniilidis, and
  Kaushik Roy.
\newblock {\em Spike-FlowNet: Event-Based Optical Flow Estimation with
  Energy-Efficient Hybrid Neural Networks}, pages 366--382.
\newblock Oct 2020.

\bibitem{Liu_2019_SelFlow}
Pengpeng Liu, Michael Lyu, Irwin King, and Jia Xu.
\newblock Selflow: Self-supervised learning of optical flow.
\newblock In {\em Proceedings of the IEEE/CVF Conference on Computer Vision and
  Pattern Recognition (CVPR)}, Jun 2019.

\bibitem{count_encoding}
Ana~I. Maqueda, Antonio Loquercio, Guillermo Gallego, Narciso Garcia, and
  Davide Scaramuzza.
\newblock Event-based vision meets deep learning on steering prediction for
  self-driving cars.
\newblock In {\em 2018 {IEEE}/{CVF} Conference on Computer Vision and Pattern
  Recognition}. {IEEE}, Jun 2018.

\bibitem{neftci_surrogate_2019}
E.~O. Neftci, H. Mostafa, and F. Zenke.
\newblock Surrogate gradient learning in spiking neural networks: {Bringing}
  the power of gradient-based optimization to spiking neural networks.
\newblock {\em IEEE Signal Processing Magazine}, 36(6):51--63, Nov 2019.

\bibitem{loihi2_2021}
Garrick Orchard, E.~Paxon Frady, Daniel Ben~Dayan Rubin, Sophia Sanborn,
  Sumit~Bam Shrestha, Friedrich~T. Sommer, and Mike Davies.
\newblock Efficient {Neuromorphic} {Signal} {Processing} with {Loihi} 2.
\newblock {\em 2021 IEEE Workshop on Signal Processing Systems (SiPS)}, pages
  254--259, 2021.

\bibitem{Scheerlinck_2020_FireNet}
Cedric Scheerlinck, Henri Rebecq, Daniel Gehrig, Nick Barnes, Robert Mahony,
  and Davide Scaramuzza.
\newblock Fast image reconstruction with an event camera.
\newblock In {\em Proceedings of the IEEE/CVF Winter Conference on Applications
  of Computer Vision (WACV)}, Mar 2020.

\bibitem{Shiba_2022}
Shintaro Shiba, Yoshimitsu Aoki, and Guillermo Gallego.
\newblock Fast event-based optical flow estimation by triplet matching.
\newblock {\em {IEEE} Signal Processing Letters}, 29:2712--2716, 2022.

\bibitem{detection_robotics}
A. Talukder and L. Matthies.
\newblock Real-time detection of moving objects from moving vehicles using
  dense stereo and optical flow.
\newblock In {\em 2004 IEEE/RSJ International Conference on Intelligent Robots
  and Systems (IROS) (IEEE Cat. No.04CH37566)}, volume~4, pages 3718--3725,
  2004.

\bibitem{teed2020raft}
Zachary Teed and Jia Deng.
\newblock Raft: Recurrent all-pairs field transforms for optical flow.
\newblock In {\em Computer Vision--ECCV 2020: 16th European Conference,
  Glasgow, UK, August 23--28, 2020, Proceedings, Part II 16}, pages 402--419.
  Springer, 2020.

\bibitem{time_lens}
S. Tulyakov, D. Gehrig, S. Georgoulis, J. Erbach, M. Gehrig, Y. Li, and D.
  Scaramuzza.
\newblock Time {Lens}: Event-based video frame interpolation.
\newblock In {\em 2021 IEEE/CVF Conference on Computer Vision and Pattern
  Recognition (CVPR)}, pages 16150--16159, Los Alamitos, CA, USA, Jun 2021.

\bibitem{stan_nature}
Stanisław Woźniak, Angeliki Pantazi, Thomas Bohnstingl, and Evangelos
  Eleftheriou.
\newblock Deep learning incorporating biologically inspired neural dynamics and
  in-memory computing.
\newblock {\em Nature Machine Intelligence}, 2:325--336, Jun 2020.

\bibitem{SLAM_Zhang}
Tianwei Zhang, Huayan Zhang, Yang Li, Yoshihiko Nakamura, and Lei Zhang.
\newblock Flowfusion: Dynamic dense {RGB-D} {SLAM} based on optical flow.
\newblock In {\em 2020 IEEE International Conference on Robotics and Automation
  (ICRA)}, pages 7322--7328, 2020.

\bibitem{Zhu_EVFlownet}
Alex Zhu, Liangzhe Yuan, Kenneth Chaney, and Kostas Daniilidis.
\newblock {EV}-{FlowNet}: Self-supervised optical flow estimation for
  event-based cameras.
\newblock In {\em Robotics: Science and Systems {XIV}}. Robotics: Science and
  Systems Foundation, Jun 2018.

\bibitem{Zhu_2018_mvsec}
Alex~Zihao Zhu, Dinesh Thakur, Tolga Ozaslan, Bernd Pfrommer, Vijay Kumar, and
  Kostas Daniilidis.
\newblock The {Multivehicle} {Stereo} {Event} {Camera} {Dataset}: An event
  camera dataset for {3D} perception.
\newblock {\em {IEEE} Robotics and Automation Letters}, 3(3):2032--2039, Jul
  2018.

\bibitem{voxel_encoding_IWE}
Alex~Zihao Zhu, Liangzhe Yuan, Kenneth Chaney, and Kostas Daniilidis.
\newblock Unsupervised event-based learning of optical flow, depth and
  egomotion.
\newblock In {\em 2019 IEEE/CVF Conference on Computer Vision and Pattern
  Recognition Workshops (CVPRW)}, pages 1694--1694, 2019.

\end{thebibliography}


\begin{thebibliography}{1}\itemsep=-1pt

\bibitem{TU_Delft}
Jesse Hagenaars, Federico Paredes-Vall\'es, and Guido de Croon.
\newblock Self-supervised learning of event-based optical flow with spiking
  neural networks.
\newblock {\em Advances in Neural Information Processing Systems}, 34, 2021.

\bibitem{spike_flownet}
Chankyu Lee, Adarsh Kosta, Alex Zhu, Kenneth Chaney, Kostas Daniilidis, and
  Kaushik Roy.
\newblock {\em Spike-FlowNet: Event-Based Optical Flow Estimation with
  Energy-Efficient Hybrid Neural Networks}, pages 366--382.
\newblock Oct 2020.

\bibitem{ParedesValls2020BackTE}
F. Paredes-Vall{\'e}s and Guido~C.H.E. de Croon.
\newblock Back to event basics: Self-supervised learning of image
  reconstruction for event cameras via photometric constancy.
\newblock {\em 2021 IEEE/CVF Conference on Computer Vision and Pattern
  Recognition (CVPR)}, pages 3445--3454, 2020.

\bibitem{Zhu_EVFlownet}
Alex Zhu, Liangzhe Yuan, Kenneth Chaney, and Kostas Daniilidis.
\newblock {EV}-{FlowNet}: Self-supervised optical flow estimation for
  event-based cameras.
\newblock In {\em Robotics: Science and Systems {XIV}}. Robotics: Science and
  Systems Foundation, Jun 2018.

\bibitem{voxel_encoding_IWE}
Alex~Zihao Zhu, Liangzhe Yuan, Kenneth Chaney, and Kostas Daniilidis.
\newblock Unsupervised event-based learning of optical flow, depth and
  egomotion.
\newblock In {\em 2019 IEEE/CVF Conference on Computer Vision and Pattern
  Recognition Workshops (CVPRW)}, pages 1694--1694, 2019.

\end{thebibliography}
}

\end{document}


\title{Supplementary Notes for\\Neuromorphic Optical Flow and Real-time Implementation with Event Cameras}


\author{Yannick Schnider$^{1,2}$, Stanisław Woźniak$^{1}$, Mathias Gehrig$^{3}$, Jules Lecomte$^{4}$, Axel von Arnim$^{4}$, 
\\Luca Benini$^{2,5}$, Davide Scaramuzza$^{3}$, Angeliki Pantazi$^{1}$
\\[4pt]
$^{1}$IBM Research -- Zurich\ \ 
$^{2}$ETH Zurich\ \ 
$^{3}$University of Zurich\ \ 
$^{4}$fortiss GmbH\ \ 
$^{5}$Università di Bologna
}

\maketitle


\newcommand{\tabBig}[1]{
\newcommand{\pout}[0]{\%\textsubscript{Out.}} 
\newcommand{\ra}[1]
{\renewcommand{\arraystretch}{#1}}
\begin{table*}[h!]
\centering
\ra{1.3}
\begin{tabular}{@{}lrrrcrrrcrrrcrrr@{}}\toprule[1.5pt]
& \multicolumn{2}{c}{outdoor\textunderscore day1} & \phantom{}& \multicolumn{2}{c}{indoor\textunderscore flying1} &
\phantom{} & \multicolumn{2}{c}{indoor\textunderscore flying2} &
\phantom{} & \multicolumn{2}{c}{indoor\textunderscore flying3} &\phantom{} & \multicolumn{2}{c}{overall}\\ \cmidrule{2-3} \cmidrule{5-6} \cmidrule{8-9} \cmidrule{11-12} \cmidrule{14-15}

$\text{dt}=1$& AEE & \pout && AEE & \pout  && AEE & \pout && AEE & \pout & \multicolumn{2}{r}{WAEE} &$\overline{\text{\%}}\textsubscript{\text{Out.}}$\\
\midrule
LIF-EV-FlowNet\cite{TU_Delft} & 0.53 & 0.33  && 0.71 & 1.41 && 1.44 & 12.75 && 1.16 & 9.11 && 0.93 & 5.90\\
XLIF-EV-FlowNet\cite{TU_Delft} & 0.45 & \textbf{0.16}  && 0.73 & \underline{0.92} && 1.45 & 12.18 && 1.17 & 8.35 && 0.90 & 5.40\\ 
LIF-FireNet\cite{TU_Delft} & 0.57 & 0.40  && 0.98 & 2.48 && 1.77 & 16.40 && 1.50 & 12.81 && 1.15 & 8.02\\
PLIF-FireNet\cite{TU_Delft} & 0.56 & 0.38  && 0.90 & 1.93 && 1.67 & 14.47 && 1.41 & 11.17 && 1.10 & 7.00\\
our SNN-Timelens & \underline{0.44} & 0.18 && \underline{0.70} & \textbf{0.79} && \underline{1.30} & \underline{9.41} && \underline{1.05} & \underline{6.00} && \underline{0.84} & \underline{4.10}\\ 
our SNUo-Timelens & \textbf{0.39} & \underline{0.17} && \textbf{0.64} & 0.96 && \textbf{1.17} & \textbf{7.71} && \textbf{0.96} & \textbf{4.92} && \textbf{0.76} & \textbf{3.44} \\\midrule 

EV-FlowNet$_{\text{PM}}$\cite{ParedesValls2020BackTE} & 0.92 & 5.4 && 0.79 & 1.2 && 1.40 & 10.9 && 1.18 & 7.4 && 1.13 &6.23 \\ 

EV-FlowNet\cite{TU_Delft} & 0.47 & 0.25 && \underline{0.60} & \underline{0.51} && \underline{1.17} & \underline{8.06} && \underline{0.93} & 5.64 && 0.78 & \underline{3.61} \\
RNN-EV-FlowNet\cite{TU_Delft} & 0.56 & 1.09 && 0.62 & 0.97 && 1.20 & 8.82 && \underline{0.93} & \underline{5.51} && 0.83 &4.10 \\ 
our sSNU-Timelens & \underline{0.36} & \underline{0.10} && \textbf{0.58} & 0.56 && 1.19 & 8.78 && 0.96 & 6.11 && \underline{0.73} & 3.89\rule[-1.2ex]{0pt}{0pt}\\ \hdashline 
\rule{0pt}{2.5ex}EV-FlowNet$_{\text{PM-MVSEC}}$ \cite{Zhu_EVFlownet} & 0.49 & 0.20 && 1.03 & 2.20 && 1.72 & 15.10 && 1.53 & 11.90 && 1.13 &7.35\\
EV-FlowNet$_{\text{CM-MVSEC}}$ \cite{voxel_encoding_IWE} & \textbf{0.32} & \textbf{0.00} && \textbf{0.58} & \textbf{0.00} && \textbf{1.02} & \textbf{4.00} && \textbf{0.87} & \textbf{3.00} && \textbf{0.67} &\textbf{1.75} \\
Hybrid-EV-FlowNet$_{\text{MVSEC}}$ \cite{spike_flownet} & 0.49 & - && 0.84 & - && 1.28 & - && 1.11 & - && 0.92 & -\\
\midrule 
\\ 
$\text{dt}=4$\\ \midrule
LIF-EV-FlowNet\cite{TU_Delft} & 2.02 & 18.91  && 2.63 & 29.55 && 4.93 & 51.10 && 3.88 & 41.49 && 0.92 & 35.26\\
XLIF-EV-FlowNet\cite{TU_Delft} & 1.67 & 12.69  && 2.72 & 31.69 && 4.93 & 51.36 && 3.91 & 42.52 && 0.89 & 34.57\\ 
LIF-FireNet\cite{TU_Delft} & 2.12 & 21.00  && 3.72 & 48.27 && 6.27 & 64.16 && 5.23 & 58.43 && 1.17 & 47.97\\
PLIF-FireNet\cite{TU_Delft} & 2.11 & 20.64  && 3.44 & 44.02 && 5.94 & 64.02 && 4.98 & 57.53 && 1.11 & 46.55\\
our SNN-Timelens & \underline{1.65} & \underline{11.03} && \underline{2.61} & \underline{29.40} && \underline{4.50} & \underline{50.87} && \underline{3.58} & \underline{40.22} && \underline{0.84} & \underline{32.88}\\  

our SNUo-Timelens & \textbf{1.44} & \textbf{8.98} && \textbf{2.36} & \textbf{24.18} && \textbf{3.98} & \textbf{44.71} && \textbf{3.25} & \textbf{36.01} && \textbf{0.75} & \textbf{28.47} \\
\midrule
EV-FlowNet\cite{TU_Delft} & 1.69 & 12.50 && \underline{2.16} & \underline{21.51} && 3.90 & \textbf{40.72} && \textbf{3.00} & \textbf{29.60} && 0.74 & \underline{26.08}\\
RNN-EV-FlowNet\cite{TU_Delft} & 1.91 &16.39 && 2.23 &22.10 && 4.01 & 41.74 &&\underline{3.07} & \underline{30.87} && 0.78 & 27.78 \\ 
our sSNU-Timelens  & 1.34 & \underline{7.99} && \textbf{2.15} & \textbf{20.92} && 3.97 & \underline{41.31} && 3.17 & 32.44 && \underline{0.71} & \textbf{25.67}\rule[-1.2ex]{0pt}{0pt}\\ \hdashline 
\rule{0pt}{2.5ex}EV-FlowNet$_{\text{PM-MVSEC}}$ \cite{Zhu_EVFlownet} & \underline{1.23} & \textbf{7.30} && 2.25 & 24.70 && 4.05 & 45.30 && 3.45 & 39.70 && 0.73 &29.25\\
EV-FlowNet$_{\text{CM-MVSEC}}$ \cite{voxel_encoding_IWE} & 1.30 & 9.70 && 2.18 & 24.20 && \underline{3.85} & 46.80 && 3.18 & 47.80 && \underline{0.71} &32.13 \\
Hybrid-EV-FlowNet$_{\text{MVSEC}}$ \cite{spike_flownet} & \textbf{1.09} & - && 2.24 & - && \textbf{3.83} & - && 3.18 & - && \textbf{0.68} & -\\ 
\bottomrule[1.5pt]   


\end{tabular}
\caption{Extended evaluation on MVSEC: AEE (the lower, the better $\downarrow$), the percentage of outliers $\text{\%}\textsubscript{\text{Out.}}$($\downarrow$) per sequence, and the overall WAEE($\downarrow$) as defined in Eq.~\ref{eq:WAEE} as well as the average percentage of outliers $\overline{\text{\%}}\textsubscript{\text{Out.}}$($\downarrow$). Best scores are in bold, while runner-ups are underlined. Horizontal lines delimit the spiking and the non-spiking models. Dashed line delimits not directly comparable prior art setups.}
\label{tab:results}
\end{table*}}


\tabBig{}

\section{WAEE metric definiton}
\label{Appendix:WAEE}
The weighted average end point error (WAEE) combines the AEEs of the four sequences:
\begin{itemize}
    \item outdoor\_day1 ($\text{od1}$)
    \item indoor\_flying1 ($\text{if1}$)
    \item indoor\_flying2 ($\text{if2}$)
    \item indoor\_flying3 ($\text{if3}$)
\end{itemize}
and is defined as: 
\begin{equation}
\begin{aligned}
  \text{WAEE} =& (\frac{\text{AEE}_{\text{od1}}}{w_{\text{od1}}} + \frac{\text{AEE}_{\text{if1}}}{w_{\text{if1}}} +\frac{\text{AEE}_{\text{if2}}}{w_{\text{if2}}}  +\frac{\text{AEE}_{\text{if3}}}{w_{\text{if3}}} )/4.\\ \nonumber
  \label{eq:WAEE}
\end{aligned}
\end{equation}

The weights $w_{\text{od1}}$, $w_{\text{if1}}$, $w_{\text{if2}}$, $w_{\text{if3}}$ are the sequence specific average AEEs of the spiking EV-FlowNet variants: LIF, ALIF, PLIF and XLIF for the modes $\text{dt} = 1$ and $\text{dt} = 4$ published in \cite{TU_Delft}:
\begin{equation}
\begin{aligned}
    \text{dt} = 1: \\
    w_{\text{od1}} =\quad &(0.53 + 0.57 + 0.60 + 0.45)/4.0 &= 0.5375 \\
    w_{\text{if1}} =\quad &(0.71 + 1.00 + 0.75 + 0.73)/4.0 &= 0.7975 \\
    w_{\text{if2}} =\quad &(1.44 + 1.78 + 1.52 + 1.45)/4.0 &= 1.5475 \\
    w_{\text{if3}} =\quad &(1.16 + 1.55 + 1.23 + 1.17)/4.0 &= 1.2775\\
    \\
    \text{dt} = 4: \\
    w_{\text{od1}} =\quad &(2.02 + 2.13 + 2.24 + 1.67)/4.0 &= 2.0150 \\
    w_{\text{if1}} =\quad &(2.63 + 3.81 + 2.80 + 2.72)/4.0 &= 2.9900 \\
    w_{\text{if2}} =\quad &(4.93 + 6.40 + 5.21 + 4.93)/4.0 &= 5.3675 \\
    w_{\text{if3}} =\quad &(3.88 + 5.53 + 4.12 + 3.91)/4.0 &= 4.3600 \\
  \label{eq:weights_WAEE}
\end{aligned}
\end{equation}

\section{Additional comparison}
Table \ref{tab:results} includes an extended comparison with additional prior art non-spiking models.
In particular, EV-FlowNet$_{\text{PM}}$ \cite{ParedesValls2020BackTE} was trained in comparable setting to ours, but used a photometric loss (PM). The results had been only reported for $\text{dt}=1$ mode.
Furthermore, several prior art architectures were trained in a different setup using directly the MVSEC dataset, as opposed to our architectures that were trained on the UZH-FPV Drone Racing Dataset and evaluated on the MVSEC dataset. Results for models trained directly on MVSEC, delimited by dashed lines, include:
\begin{itemize}
    \item EV-FlowNet$_{\text{PM-MVSEC}}$ \cite{Zhu_EVFlownet}, trained in a self-supervised manner with the photometric loss (PM),
    \item EV-FlowNet$_{\text{CM-MVSEC}}$ \cite{voxel_encoding_IWE}, trained in a self-supervised manner with a contrast maximisation loss (CM),
    \item Hybrid-EV-FlowNet$_{\text{MVSEC}}$ \cite{spike_flownet}, trained in a self-supervised manner with the photometric loss.
\end{itemize}

Considering the extended comparison with non-spiking ANN prior art models, the EV-FlowNet$_{\text{CM-MVSEC}}$ \cite{voxel_encoding_IWE} yields the best performance on all MVSEC sequences for $\text{dt} = 1$ with regard to WAEE and percentage of outliers. Its WAEE of 0.67 is 8.2\% lower than 0.73 of our sSNU-Timelens. In turn, the Hybrid-EV-FlowNet$_{\text{MVSEC}}$ \cite{spike_flownet} is outperformed by our sSNU-Timelens by 26.0\% (0.73 vs. 0.92). 

However, when evaluating in mode $\text{dt} = 4$, the Hybrid-EV-FlowNet$_{\text{MVSEC}}$ \cite{spike_flownet} yields the best overall performance with an WAEE of 0.68 compared to our sSNU-Timelens with 0.71 (+4.4\%). The sSNU-Timelens shows on par performance in terms of WAEE with the EV-FlowNet$_{\text{CM-MVSEC}}$ \cite{voxel_encoding_IWE} (also 0.71) in this mode. 

In summary, the EV-FlowNet$_{\text{CM-MVSEC}}$ \cite{voxel_encoding_IWE} and the Hybrid-EV-FlowNet$_{\text{MVSEC}}$ \cite{spike_flownet} perform best for MVSEC evaluations with $\text{dt} = 1$ and $\text{dt} = 4$, respectively. Remarkably, our sSNU-Timelens is a runner-up in both cases, despite being trained without access to the examples from the MVSEC dataset.




{\small
\bibliographystyle{ieee_fullname}
\bibliography{egbib}
}